\documentclass[11pt,a4paper]{article}
\usepackage[hyperref]{acl2020}
\usepackage{times}
\usepackage{latexsym}
\usepackage{graphicx}
\usepackage{multirow}
\usepackage{booktabs}

\usepackage{microtype}

\aclfinalcopy 

\title{What does BERT Learn from Multiple-Choice \\ Reading Comprehension Datasets?}

\author{Chenglei Si$^1$, Shuohang Wang$^2$, Min-Yen Kan$^3$, Jing Jiang$^4$ \\
  $^1$River Valley High School, $^2$Microsoft Dynamics 365 AI Research, \\ $^3$School of Computing, National University of Singapore\\
  $^4$Singapore Management University \\
   {\tt sichenglei1125@gmail.com, shuowa@microsoft.com} \\
  {\tt kanmy@comp.nus.edu.sg, jingjiang@smu.edu.sg}
}

\begin{document}
\maketitle
\begin{abstract}
Multiple-Choice Reading Comprehension (MCRC) requires the model to read the passage and question, and select the correct answer among the given options. Recent state-of-the-art models have achieved impressive  performance on multiple MCRC datasets. However, such performance may not reflect the model's true ability of language understanding and reasoning. 
In this work, we adopt two approaches to investigate what BERT learns from MCRC datasets: 1) an \textbf{un-readable data attack}, in which we add keywords to confuse BERT, leading to a significant performance drop; and 2) an \textbf{un-answerable data training}, in which we train BERT on partial or shuffled input. Under un-answerable data training, BERT achieves unexpectedly high performance. 
Based on our experiments on the 5 key MCRC datasets --- RACE, MCTest, MCScript, MCScript2.0, DREAM --- we observe that 1) fine-tuned BERT mainly learns how keywords lead to correct prediction, instead of learning semantic understanding and reasoning; and 2) BERT does not need correct syntactic information to solve the task; 3)
there exists artifacts in these datasets such that they can be solved even without the full context.
\end{abstract}

\section{Introduction}
The pre-trained language models, such as ELMo~\cite{ELMO}, GPT~\cite{GPT}, BERT~\cite{BERT}, XLNet~\cite{xlnet}, RoBERTa~\cite{RoBERTa}, ALBERT~\cite{ALBERT} have drawn much attention recently as they have achieved the state-of-the-art on a wide spectrum of NLP tasks.
So far the Tranformer~\cite{transformer} based model, BERT, is the most widely adopted baseline on different tasks, such as Machine Reading Comprehension, Text Entailment, Sentiment Analysis, Relation Extraction, Dependency Parsing, and many others~\cite{BERT,tuneOrNot,transferability}. 
How it achieves its performance is of paramount importance in guiding NLP research to rectify its model weakness and further leverage its advantages.

Due to the high complexity of deep neural network, it is still hard to explain how it works by direct analysis of its parameters.
Most analysis works treat it as a black-box and use external probes to assess the robustness and weakness of the models. 
We follow this black-box paradigm and design new probing tasks to explore what attributes to the high performance of BERT. In this work, we focus on the task of Multiple-Choice Reading Comprehension (MCRC), which requires the model to select the correct answer among several candidate options, given a passage and question. Our probing approach is to attack the fine-tuned BERT by adding distracting information with keywords during testing.  
If the performance under attack significantly drops, we can infer that BERT relies too heavily on keyword matching.

For this approach, which we term the \textbf{Un-Readable Data Attack}, we explore three adversarial attack methods by adding additional distracting information into either the passage or the original distractors to test the model's robustness. 
The goal of our adversarial attack is to manipulate the input to mislead the model into making incorrect predictions, while humans will still be able to choose the correct answers under these attacks. 
Our distracting information is mainly in the form of un-readable (i.e. uninterpretable) sentences which are created by shuffling the word order in the original input, as shown in the example in Figure~\ref{AddSent2Pas_example}.
Based on our experiment results, we find that BERT is easily distracted by the un-readable information, which suggests that it heavily relies on the statistical patterns such as keyword matching to achieve high performance.


    \begin{figure}[t]
      \small
      \framebox{\parbox{7.5cm}{
      \textbf{Source: MCScript}
      
      \textbf{Passage:}
      Early this morning, I woke up to the sound of my newspaper landing on my driveway. I sat up and wrapped my pink robe around me. I slipped my feet into my slippers and looked at the clock. It was only 7:00 but it was time for me to get my newspaper and drink some coffee. I looked out the window and noticed it was raining quite a bit ......
       \textit{ \textbf{(Appended Adversary Sentence:) time in they 00 the wake What am morning up : 9?}}
      
      \textbf{Question:} What time they wake up in the morning?
      
      \textbf{Options:} 
      
      A (Distractor). 9:00 am
      
      B (Answer). seven o'clock 
      
      
}}
      \caption{An example of Multi-Choice Reading Comprehension task (MCRC). The last sentence of the passage in bold is the un-readable sequence we append to attack BERT.}
      \label{AddSent2Pas_example}
 \end{figure}
 

While the MCRC task has been gaining intense interest among the research community, it remains unclear as of what information is necessary for the model to achieve high performance on these datasets. 
To investigate this problem, we fine-tune BERT on the questions where humans can only randomly guess the answer, but where keywords remain in the training set. 

For this direction, which we call \textbf{Un-Answerable Data Training}, we first try partial training where we remove the passage, the question, or both,  when training the model.
 Next, we train BERT with shuffled input, where we randomly shuffle all the words in the passage, or question, or both. Under both settings, humans would not be able to learn anything from the training input. 
However, we observe that the performance of BERT trained under these two settings is much better than a random guess baseline for all the 5 MCRC test sets.
This suggests that BERT does not need correct syntactic information to answer the questions and there exist artifacts and statistical cues within all these datasets so that BERT can perform well even without enough context.

We summarise our main contributions in this work:

\begin{itemize}
    \item We are the first to conduct a deep analysis of the SOTA model, BERT, on MCRC datasets.
    \item We propose 3 methods to construct un-readable data to attack BERT on MCRC datasets and these methods will make the performance significantly drop.
    \item We find that using un-answerable data to train BERT on MCRC can still achieve good performance.
    \item Based on all the experiment results from two different aspects, we observe that BERT mainly learns the key statistical patterns for selecting the answer instead of semantic understanding; BERT can solve the task without the correct word order; and current benchmark datasets do not truly test the model's ability of language understanding. 
        
    
    
    
    
    

\end{itemize}

\section{Related Work}
The interpretability and analysis of models for NLP tasks usually fall into three directions: adversarial attacks to reveal the weaknesses of NLP models, partial data training to assess the quality of datasets, and downstream tasks to test the linguistic properties of the model. 

\textbf{Adversarial attacks} in NLP have attracted increased interest in recent years and have been explored on several related tasks. \citet{AdvSquad} show that by appending a distractor sentence generated by manually defined rules, the state-of-the-art performance on the SQuAD dataset drops significantly.  \citet{SEA} systematically generate semantic equivalent adversaries by paraphrasing the questions. 
\citet{hotflip} generate adversaries by replacing characters or words in the input sequence based on the gradient of the one-hot input vectors. \citet{GenAdv} develop a black-box attack algorithm that exploits population-based gradient-free optimization via genetic algorithms.  
\citet{SCPN}  propose syntactically controlled paraphrase networks and use them to generate adversarial examples that follow the target syntactic form. 
\citet{universalTrigger} use a gradient-guided search to find universal adversarial sequences that trigger a model to produce a specific prediction when concatenated to any input from a dataset. \citet{shuffleDialog}
observe that neural dialog architectures models are insensitive to most sequence perturbations.



\textbf{Partial data training} has also been adopted to test the model and the datasets, such as machine comprehension  and natural language inference tasks.  \citet{reading} find that question- and passage-only models perform surprisingly well on extractive machine comprehension datasets, which suggests that some datasets may not be challenging enough. 
\citet{hypoBasline} and  \citet{artifacts} find that a hypothesis-only model is able to significantly outperform a majority class baseline across a number of NLI datasets and they attribute this to statistical irregularities and annotation artifacts of the datasets. 
\citet{statsCue} probe BERT training by removing either the warrants, claims or reasons from the  Argument Reasoning
Comprehension Task and observe only a small decrease of performance compared to the full training setting, thus revealing that BERT relies heavily on spurious statistical cues.

\textbf{Analysis by downstream tasks} is also a recent line of work to analyze the linguistic properties of NLP models.  
\citet{BERTsyntax} find that BERT performs very well on subject-verb agreement tasks, showing its sensitivity to syntax. 
By probing the attention heads of BERT,
\citet{BERTattn} find that  certain attention heads correspond well to linguistic notions of syntax and coreference, and that substantial syntactic information is captured in
BERT’s attention. 
\citet{transferability} use seventeen diverse probing tasks and observe that linear models trained on top of frozen contextual representations are competitive with state-of-the-art
task-specific models in many cases, but fail on
tasks requiring fine-grained linguistic knowledge.
\citet{BERTpipeline} find that BERT represents the steps of the traditional NLP pipeline in an interpretable and localizable way.
\citet{BERT-NPI} test BERT on negative polarity item (NPI) licensing in English, and find that BERT has significant knowledge of these grammatical features.

In this work, we mainly focus on the first two directions. We propose three new types of adversarial attack methods based on shuffling, which is simple and effective to test the robustness of the model.
Moreover, we extend the partial data training method to shuffled training. Instead of removing the whole passage or question, our proposed method is to shuffle them to make the sequences uninterpretable while keeping all the keywords. It serves to analyse what kind information is required to solve MCRC datasets. 

\section{Un-Readable Data Attack} \label{sec: attack_sec}
\begin{figure*}[h]
    \centering
    \includegraphics[width=15.5cm]{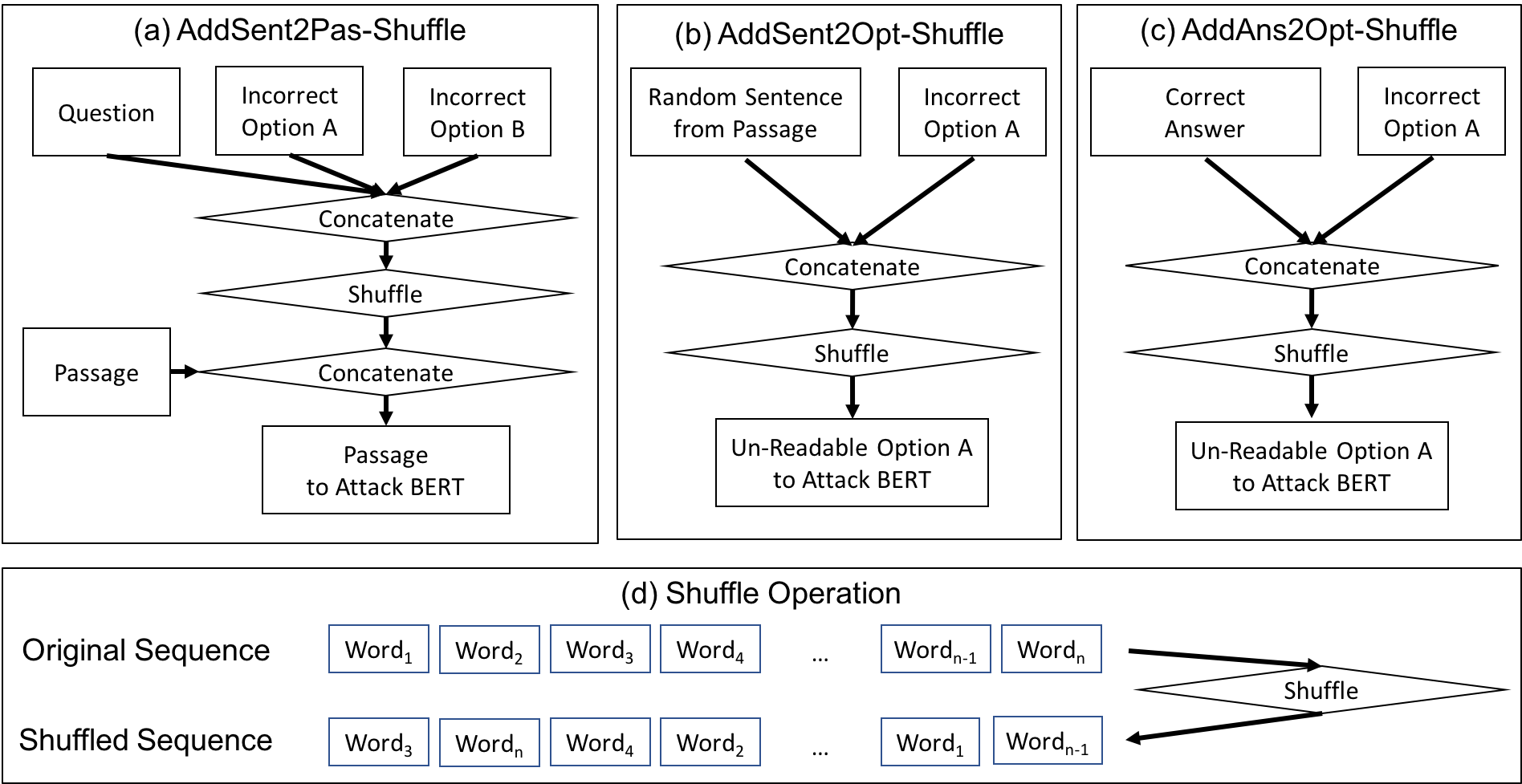}
    \caption{The procedures of constructing un-readable data to attack BERT.}
    \label{attack_diagram}
\end{figure*}

In this section, we introduce the methods of constructing un-readable data to attack BERT. We first fine-tune BERT on the original MCRC data and then test it under adversarial attacks.
The un-readable data is mainly obtained by randomly shuffling the word order of the input to make it grammatically wrong and un-readable.
Note that we do not shuffle the correct answers but only shuffle the constructed distractors, and hence after adding the un-readable distractors to the original MCRC test data, the labels will remain the same.
We propose three methods of using un-readable data to attack BERT and investigate what BERT actually learns. An overview of our attack methods are shown in Figure~\ref{attack_diagram} and we will explain each one of them in detail in this section.

\subsection{Un-Readable Sub-Passage}
\paragraph{AddSent2Pas-Shuffle} One way of adversarial attack is to add the un-readable data to the passage.
An example is shown in Figure~\ref{AddSent2Pas_example}.
Inspired by the ``AddSent" method~\cite{AdvSquad}, we append an un-readable sequence to the end of the passage to distract BERT.
The sequence consists of the question and all the original distractors. 
Note that directly appending the raw sequence to the passage, we may make some incorrect options correct.
To avoid such confusion, we randomly shuffle the sequence to make it un-readable with the following restriction on \textit{ShuffleDegree}:  
$$ShuffleDegree = \frac{MinimumEditDistance}{SequenceLength}$$
The Minimum Edit Distance (MED) is computed between the original sequence and the shuffled sequence.
We set a threshold for the $ShuffleDegree \in [0,1]$ and reshuffle the sequence until the \textit{ShuffleDegree} is higher than the threshold. \footnote{Empirically we find that for all of our attack methods, the average \textit{ShuffleDegree} on all the datasets is above 0.65, which is sufficient to make the sequence ungrammatical and hard to interpret.}
We also reshuffle the sequence if there is any exact match between the incorrect options and a subsequence of the shuffled sequence.
This ensures that only meaningless information is added to the passage and the answer of the question does not change.
The main difference triggered by our method is that some keywords from the incorrect options are now present in the passage.
Ideally, if BERT can fully understand the text, it will not be fooled by the appended un-readable sequence. 
If it is fooled, we could conclude that BERT relies heavily on some statistical patterns, like the passage-option keywords matching, from the training data.

\subsection{Un-Readable Options}
\label{sec:unreadable}
Another way of attacking BERT is to add the un-readable data to the original distractors of the questions to form the new distratcors and use these constructed new distractors to replace the original distractors. We propose two methods to construct such attack.

\paragraph{AddSent2Opt-Shuffle}
In this method, for each original distratcor of the question, we randomly select one sentence from the passage and append it to the original distractor. 
After appending, the new distractors remain wrong, as the wrong facts from the original distratcors are not removed.
To make the new distratcor un-readable, we further shuffle these new distrators. The threshold of  for the \textit{ShuffleDegree} restriction is also imposed for shuffling.
Note that, the correct answer is not modified and we only make the other incorrect options un-readable.
In this way, it should be very easy for humans to select the correct answer since the distractors are now uninterpretable and easy to be distinguished between the correct answer.
If BERT gets much worse performance on these new options, it means that BERT relies too much on the keywords matching, as the new distractors share more matched words with the passage now, even though the shuffling disrupts their original semantic and syntactic information.

To further explore the attacking method, we also compare it with another two variants,  \textbf{AddSent2Opt}, which only appends one  randomly selected sentence from the passage to the original distractor without further shuffling, and \textbf{Sent2Opt-Shuffle}, which directly shuffles the randomly selected sentence from passage to replace the original distractor.
An example is shown in Figure~\ref{AddSent_example}.


\paragraph{AddAns2Opt-Shuffle}
In this method, instead of appending one sentence selected from the passage, we directly append the correct answer to the end of each original distractor to form the new distractor options. 
The new distractor options should still be considered wrong since part of them come from the original incorrect options. 
We further shuffle these new options to make them un-readable.

Similarly, we experiment with two other attack methods for comparison: 
\textbf{AddAns2Opt}: We append the correct answer to the end of each original distractor to form the new distractors, without shuffling.
\textbf{Ans2Opt-Shuffle}: We directly shuffle the correct answer to replace the original distractors.

    \begin{figure}[t]
      \small
      \framebox{\parbox{7.5cm}{
      \textbf{Source: RACE-H}
      
      \textbf{Passage:} Mike is a factory worker. \textbf{\textit{(Randomly Selected Sentence:) He is often very tired after a day's work.}} His wife, Jenny, has no job, so she stays at home to cook the meals...
        
      \textbf{Question:} Jenny stays at home because \_ .
        
      \textbf{Original Options:} 
      
      A. she likes cooking
      
      B. she loves her husband very much
      
      C [Answer]. she doesn't have a job
      
      D. she doesn't want to work
      
      \textbf{AddSent2Opt Options:} 
     
      A. she likes cooking \textit{He is often very tired after a day's work.}
      
      B. she loves her husband very much \textit{He is often very tired after a day's work.}
      
      C [Answer]. she doesn't have a job
      
      D. she doesn't want to work \textit{He is often very tired after a day's work.}
      
      
      
      \textbf{Sent2Opt-Shuffle Options:}
      
      A. \textit{'}s work . is very He after a tired often day
      
      B. is tired often . day a \textit{'}s work after He very
      
      C [Answer]. she doesn't have a job
      
      D. is very \textit{'}s often after work tired a day He .
      
    
    \textbf{AddSent2Opt-Shuffle Options:}
    
    A. He . a often she work tired likes cooking is \textit{'}s very day after
    
    B. \textit{'}s very day very tired work her is often . much He loves husband after a she
    
    C [Answer]. she doesn't have a job
    
    D. work . often He want work to \textit{'}s very tired n\textit{'}t does after a day she is
}}
      \caption{Examples of AddSent2Opt, Sent2Opt-Shuffle and AddSent2Opt-Shuffle. The sentence in bold from the passage is randomly selected to be added to the incorrect options.}
      \label{AddSent_example}
   \end{figure}

\begin{figure}[t]
      \small
      \framebox{\parbox{7.5cm}{
      \textbf{Source: DREAM}
      
      
      
      
      \textbf{Shuffled Passage:}
      
      making not class . definitely If stick any dancing worth time : am it effort W with considering I were and . I my you am progress : . It's I M dropping , I .
        
      \textbf{Question:} What does the man suggest the woman do?
      
        
      \textbf{Options:} 
      
      A. Consult her dancing teacher.
      
      B. Take a more interesting class.
      
      C [Answer]. Continue her dancing class.
      
}}
      \caption{Example of un-answerable data to train BERT. The passage is shuffled.}
      \label{shuffle_example}
   \end{figure}
\section{Un-Answerable Data Training}
For all the methods in the previous section, we are attacking BERT which has been trained on the original MCRC dataset. 
In this section, we will train BERT with un-answerable data, from which human beings can learn little knowledge about the ways to answer questions.
We test whether under this setting, BERT could still achieve higher performance than random guess.
Note that the test set is not modified and still follows the original setting.
We will introduce two methods for constructing un-answerable data.

\subsection{Shuffled Data} 
To make the questions un-answerable, one simple way is to shuffle the word order the original text, so that the shuffled texts do not follow the correct grammar and become un-readable.
Supposedly the passage is the necessary information to answer the questions in reading comprehension tasks, the questions will be un-answerable by randomly shuffling the passage words.
Although the shuffled passage doesn't follow the grammar any more, all the original keywords are kept. 
If BERT can still achieve high performance on shuffled data, we can conclude that BERT does not need the correct syntactic information to answer the questions.

Based on this motivation, we try three different settings of shuffled training data: 1) shuffle words in the passage  (\textbf{P-Shuffle}), as shown in Figure~\ref{shuffle_example}. 2) shuffle words in the question  (\textbf{Q-Shuffle}), and 3) shuffle both the passage and the question (\textbf{PQ-Shuffle}). 


\subsection{Partial Data}
In another way, instead of shuffling either the passage or the question to construct the un-answerable data to train BERT, we directly remove all the passages or the questions in training data.
With partial information left, human beings are also impossible to learn how to select the correct answer.

Similar to the shuffled data setting, we also try three different settings: 1) remove the passage  (\textbf{P-Remove}), 2) remove the question  (\textbf{Q-Remove}), and 3) remove both the passage and the question (\textbf{PQ-Remove}). 





\section{Experiment}
In this section, we introduce the datasets we use to test BERT, our experiment results and our further analysis.
\subsection{Datasets}

\begin{table}[t]
\centering
\small 
\addtolength{\tabcolsep}{-2pt}
\begin{tabular}{ lcc }
\toprule
 & Train/Dev/Test
 & Pas/Que/Ans
 \\
 \midrule
MC160 & 280/120/240 & 204/8/3  \\
MC500 & 1200/200/600 & 212/8/3  \\
RACE-M & 25421/1436/1436 & 231/9/4  \\
RACE-H & 62445/3451/3498 & 353/10/6  \\
MCScript & 9731/1411/2797 & 196/8/4 \\
MCScript2.0 & 14191/2020/3610 & 164/8/3 \\
DREAM & 6116/2040/2041 & 86/9/5 \\
\bottomrule
\end{tabular}
\caption{Dataset Statistics.}
\label{table_data}
\end{table}

\begin{table*}[t]
\centering
\small 
\begin{tabular}{ lcccccccc }
\toprule
 & MC160
 & MC500
 & RACE-M
 & RACE-H
 & MCScript
 & MCScript2.0
 & DREAM
 & Average
 \\
\midrule 
Random Guess & 25.0 & 25.0 & 25.0 & 25.0 & 50.0 & 50.0 & 33.3 & - \\
BERT & 74.7 & 69.3 & 75.6 & 64.7 & 87.7 & 83.9 & 62.8 & - \\
\midrule

\multirow{2}{*}{AddSent2Pas-Shuffle} & {32.1} & {31.6} & {\textbf{41.0}} & {\textbf{34.5}} & {36.2} & {41.2} & {42.0} & - \\

& \textit{-57.0\%} & \textit{-54.4\%} & \textit{\textbf{-45.8\%}} & \textit{\textbf{-46.7\%}} & \textit{-58.7\%} & \textit{-50.9\%} & \textit{-33.1\%} & \textit{-49.5\%} \\



\multirow{2}{*}{AddSent2Opt-Shuffle} & 46.5 & 43.4 & 58.8 & 50.3 & 29.9 & 25.5 & 59.3 & - \\
& \textit{-37.8\%} & \textit{-37.4\%} & \textit{-22.2\%} & \textit{-22.3\%} & \textit{-65.9\%} & \textit{-69.6\%} & \textit{-5.6\%} & \textit{-37.3\%} \\

\multirow{2}{*}{AddAns2Opt-Shuffle}  & 73.5 & 66.2 & 65.1 & 50.0 & 75.4 & 65.4 & 76.1 & - \\
 
 & \textit{-1.6\%} & \textit{-4.5\%} & \textit{-13.9\%} & \textit{-22.7\%} & \textit{-14.0\%} & \textit{-22.1\%} & \textit{+21.1\%} & \textit{-8.2\%} \\
 
 


\multirow{2}{*}{Sent2Opt-Shuffle}
& {37.1} & {36.7} & {48.3} & {43.3} & {\textbf{14.5}} & {\textbf{15.4}} & {\textbf{30.6}} & - \\

& \textit{-50.3\%} & \textit{-47.0\%} & \textit{-36.1\%} & \textit{-33.1\%} & \textit{\textbf{-83.5\%}} & \textit{\textbf{-81.6\%}} & \textit{\textbf{-51.3\%}} & \textit{-54.7\%} \\

\multirow{2}{*}{Ans2Opt-Shuffle} 
& {68.8} & {63.6} & {49.1} & {44.1} & {55.6} & {52.1} & {41.2} & -  \\

& \textit{-7.9\%} & \textit{-8.2\%} & \textit{-35.1\%} & \textit{-31.8\%} & \textit{-36.6\%} & \textit{-37.9\%} & \textit{-34.4\%} & \textit{-27.4\%} \\ 

\multirow{2}{*}{AddSent2Opt} & {\textbf{17.5}} & {\textbf{19.1}} & {60.0} & {49.6} & {38.6} & {34.4} & {35.2} & - \\

& \textit{\textbf{-76.6\%}} & \textit{\textbf{-72.4\%}} & \textit{-20.6\%} & \textit{-23.3\%} & \textit{-56.0\%} & \textit{-59.0\%} & \textit{-43.9\%} & \textit{-50.3\%} \\

\multirow{2}{*}{AddAns2Opt} & {47.9} & {38.5} & {60.1} & {43.6} & {79.2} & {69.6} & {47.9} & -\\

& \textit{-35.9\%} & \textit{-44.4\%} & \textit{-20.5\%} & \textit{-32.6\%} & \textit{-9.7\%} & \textit{-17.0\%} & \textit{-23.7\%} & \textit{-26.3\%} \\

 

 



\midrule
Average Drop & \textit{-38.2\%} & \textit{-38.3\%} & \textit{-27.7\%} & \textit{-30.4\%} & \textit{-46.3\%} & \textit{-48.3\%} & \textit{-24.4\%} &  \textit{-36.2\%} \\ 





\bottomrule
\end{tabular}
\caption{Results for un-readable data attacks. Numbers in \textit{italics} are percentage change relative to the  original performance. The most effective attack method on each dataset is in \textbf{bold}.}
\label{adv_attack}
\end{table*}



\begin{table*}[t]
\centering
\small 
\begin{tabular}{ lccccccc }
\toprule 
  & MC160
 & MC500
 & RACE-M
 & RACE-H
 & MCScript
 & MCScript2.0
 & DREAM
 \\
 \midrule
Random Guess & 25.0 & 25.0 & 25.0 & 25.0 & 50.0 & 50.0 & 33.3 \\
Longest Baseline & 34.6 & 35.0 & 29.1 & 29.2 & 55.0 & 58.7 & 34.3 \\
\midrule
P-Shuffle & 60.2& 50.8 & 63.2 & 56.6 & 86.5 & 81.6 & 46.8 \\
Q-Shuffle & 70.8 & 62.9 & 72.7 & 62.5 & 86.7 & 83.6 & 50.5 \\
PQ-Shuffle & 60.8 & 49.2 & 60.6 & 55.0 & 83.3 & 77.0 & 41.2 \\
\midrule
P-Remove & 38.7 & 38.7 & 48.1 & 51.5 & 76.8 & 73.6 & 41.9 \\
Q-Remove & 61.7 & 59.5 & 57.7 & 57.8 & 84.5 & 80.2 & 62.2 \\
PQ-Remove & 31.8 & 38.3 & 41.9 & 45.3 & 72.5 & 68.1 & 41.5 \\

 \bottomrule
\end{tabular}
\caption{Results for shuffled and partial data training.}
\label{shuffle_input}
\end{table*}

In this paper we analyse 5 MCRC datasets. We briefly introduce each of them in this sub-section and provide dataset statistics, including number of questions split and average length of passage, question and answer in Table \ref{table_data}.

\noindent \textbf{MCTest (MC160 \& MC500)} \cite{MCTest} The passages in this dataset are open-domain fictional stories written by crowdsource workers.
MCTest is divided into two sets: MC160 and MC500. MC160 was gathered first, then improvements were made before gathering MC500. Each question has four options and one of them is correct.

\noindent \textbf{MCScript} \cite{MCScript} This dataset focuses on commonsense knowledge about sequences of events describing stereotypical human activities, co-called scripts. Answering a substantial subset of questions 
requires inference using commonsense knowledge about everyday activities. Each question has two options and one of them is correct.

\noindent \textbf{MCScript2.0} \cite{mcscript2}
About half of the questions require the use of commonsense and script knowledge for finding the correct answer, a notably higher number than in MCScript. In comparison to MCScript, commonsense-based questions in MCScript2.0 are harder to answer. Each question has two options and one of them is correct. The test set has not been released yet so all results in this paper are evaluated on its dev set. We trained our model on the combined training set of MCScript and MCScript2.0.

\noindent \textbf{RACE (RACE-M \& RACE-H)} \cite{RACE} Unlike other datasets where the passages are crowdsoured, the passages in this dataset are collected from the English exams for middle and high school Chinese students.
The proportion of questions that requires reasoning is larger in RACE than other MCRC datasets. It is split into RACE-M and RACE-H, which comes from middle and high school exams respectively and RACE-H is more difficult. Each question has four options and one of them is correct.

\noindent \textbf{DREAM} \cite{DREAM} This is a dialogue-based multiple-choice reading comprehension dataset, collected from English-as-a-foreign-language examinations designed by human experts to evaluate the comprehension level of Chinese learners of English, focusing on in-depth multi-turn multi-party dialogue understanding. Each question has three options and one of them is correct.


\subsection{Experiment Results}
\paragraph{Experiment setting} All of our models are based on BERT$_{LARGE}$ model,~\footnote{Our code follows the open-sourced work to finetune MCRC model,  https://github.com/huggingface/pytorch-transformers} which is a Transformer with 24 layers, 16 heads and 340M parameters in total.
During training, the passage, question and each option are concatenated as a new sequence to run BERT. 
The segment embeddings for passage words are set as 0, and the left words are 1.

\paragraph{Un-readable data attack} experiment results are shown in Table~\ref{adv_attack}.
The performance of BERT, which is finetuned on the original MCRC dataset, is shown on the top of the table.
This finetuned BERT is attacked by our different methods in Section~\ref{sec: attack_sec}, and the results are shown in the middle of the table.
Note that our attacks do not change the questions and the correct answers, and so the ground-truth of the questions is also not changed. 
All the methods named with ``Shuffle" will add un-readable information to either the passage or the distratcors.
Based on the experiment results, we can clearly see that the performance of BERT significantly drops on almost all the datasets, and for some datasets the performance is even lower than random guess.
It means that BERT is not able to detect the correct word order or the grammar, and is heavily relying on the keywords matching.
For example, BERT drops around 50\% on average on all the datasets under the attack of ``AddSent2Pas-Shuffle"'.
Although the appended data is not readable for human, it is able to fool BERT.

The methods named with ``Add'' will append additional information to the original distractors. 
 The methods without ``Add'' directly use the shuffled additional information as the new distractors to attack BERT. 
By comparing the results of ``AddSent2Opt-Shuffle" and ``Sent2Opt-Shuffle",  both of which will make the distratcors un-readable, we can see that the performance under ``AddSent2Opt-Shuffle" attack is generally better.
In this way, we may conclude that the words in the original distractors also play an important role to make BERT get the correct answer.


We also observe that different datasets suffer to a different extent to the different attack methods. 
For example, MC160 and MC500 are most sensitive to the attack method ``AddSent2Opt", which directly appends one sentence from passage to the original distractors. 
As the datasets of MC160 and MC500 are relatively small, the exact matching between the option and the passage still plays the most important role for BERT to select the answer. 
In the way, the attack can make the performance even worse than random guess. 
RACE is most sensitive to AddSent2Pas-Shuffle, while MCScript, MCScript2.0 and DREAM are most sensitive to Sent2Opt-Shuffle. 
These difference may be caused by the different natures of the datasets, such as the sources of the passages, questions and options; and the different distributions of the questions (e.g. proportion of matching, single-sentence reasoning, multiple-sentence reasoning and arithmetic questions). 


\paragraph{Un-answerable data training} experiments are shown in Table~\ref{shuffle_input}.
From the experiments, we can clearly see that BERT trained on all these settings are much better than random guess and ``Longest Baseline" - which always selects the longest option as the prediction.
For example, BERT trained under the setting of ``P-Shuffle", which randomly shuffles all the passage words in the training set, even gets very close performance to the original full training setting, especially on the MCScript and MCScript2.0 datasets. 
If we further compare ``P-Shuffle" and ``P-Remove", we can find that although the shuffled passage does not follow the correct grammar and is not interpretable any more, it is still better than removing the whole passage.
Similar performance can also be found by comparing ``Q-Shuffle" and ``Q-Remove", ``PQ-Shuffle" and ``PQ-Remove".
In this way, BERT can still learn from the shuffled input, which suggests that it is insensitive to the word order or the syntactic information. 

The experiments in Table~\ref{shuffle_input} not only reveal the behaviour of BERT,
but also reflect the general problem of all the Multi-Choice Reading Comprehension datasets themselves.
For example, according to the performance of ``Q-Remove'' and ``P-Remove'', without even reading the question or the passage, BERT can already achieve much better performance than random guess, or sometimes even close to the full training setting (for instance, BERT achieves 76.8\% accuracy on MCScript and 73.6\% accuracy on MCScript2.0 test set when trained without the passages.) 
This suggests that BERT may be exploiting dataset artifacts and statistical cues to achieve high performance.
In this way, to test the true ability of neural models to comprehend the passages, questions, options and perform reasoning, we need to be more careful when constructing the datasets.

\subsection{Further Analysis}
In this section, we analyse whether shuffle-based attack method is influenced by the answer length and the randomness of the shuffled sequence, and we will also have a case study to illustrate how the un-readable data affect the predicted probabilities of different options. 








\begin{figure}[t]
    \centering
    \includegraphics[width=7.5cm]{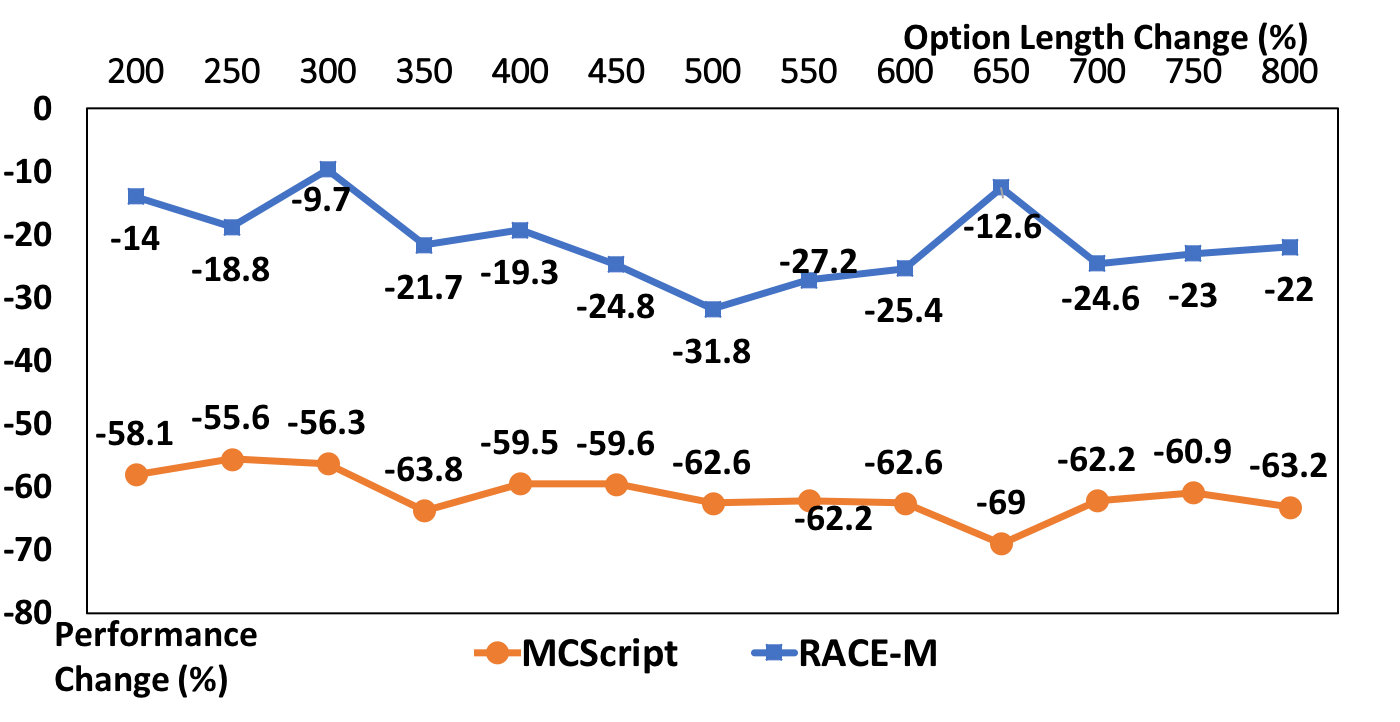}
    \caption{Plot of percentage performance change against percentage length change of the options. \\ Correlation Coefficient: MCScript: -0.6403, RACE-M: -0.3997.}
    \label{option_len_change}
\end{figure}

\begin{figure}[t]
    \centering
    \includegraphics[width=7.5cm]{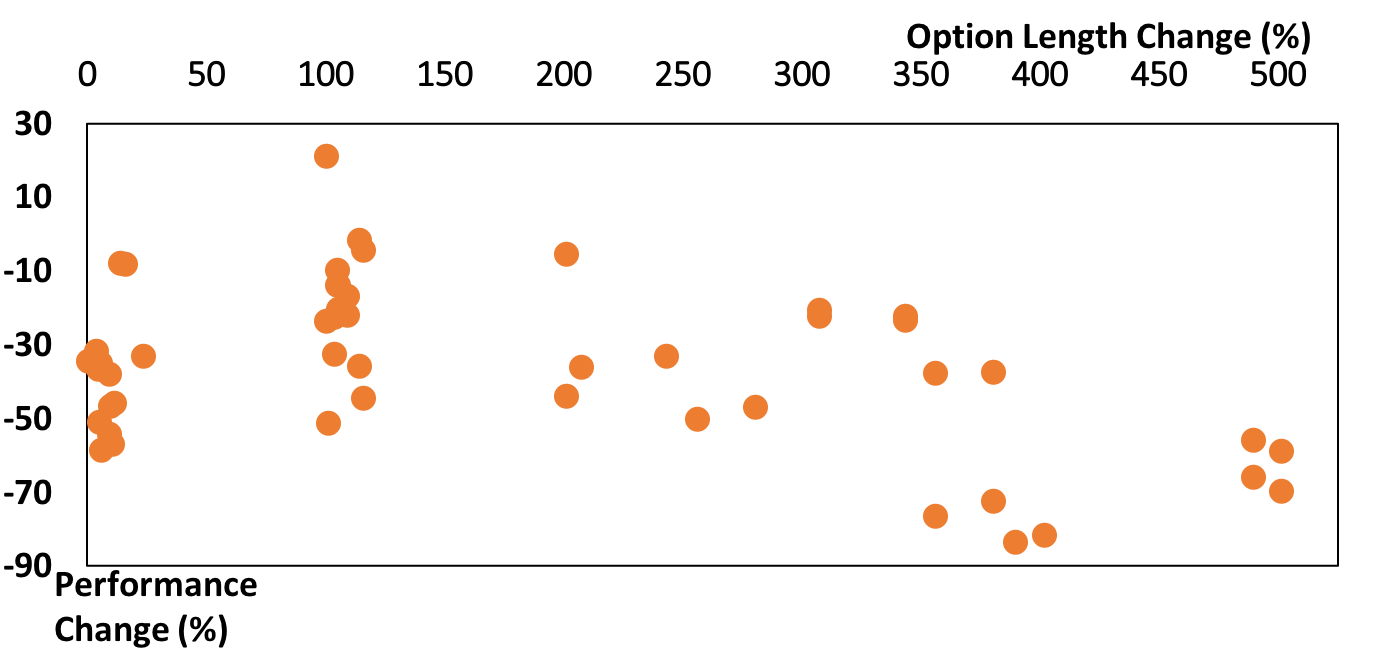}
    \caption{Plot of percentage performance change against percentage option length change, each point refers to an attack method on a dataset. Correlation Coefficient: -0.4293.}
    \label{49_points}
\end{figure}

\paragraph{Effect of attacking sequence length change} 
As our attacking method ``AddSent2Opt-Shuffle" appends additional information to the original incorrect option, we test whether the performance drop is affected by the distractor length change (constructed new distractors compared to original distractors).
We plot the percentage performance drop against percentage length change of the distratcors under ``AddSent2Opt-Shuffle" on RACE-M and MCSCript in Figure~\ref{option_len_change}.
It shows that these two factors are not correlated, with relatively large negative correlation coefficient scores on both datasets.

Furthermore, for all the attacking results in Table~\ref{adv_attack}, we compute the average  sequence length change of each attacking method on each dataset and plot all the points in Figure~\ref{49_points}. 
We also get a quite large negative correlation coefficient score.
It also shows that, in general, changing original sequence length from the datasets is not the main reason why the attacks can fool BERT.

\begin{figure}[h]
    \centering
    \includegraphics[width=7.5cm]{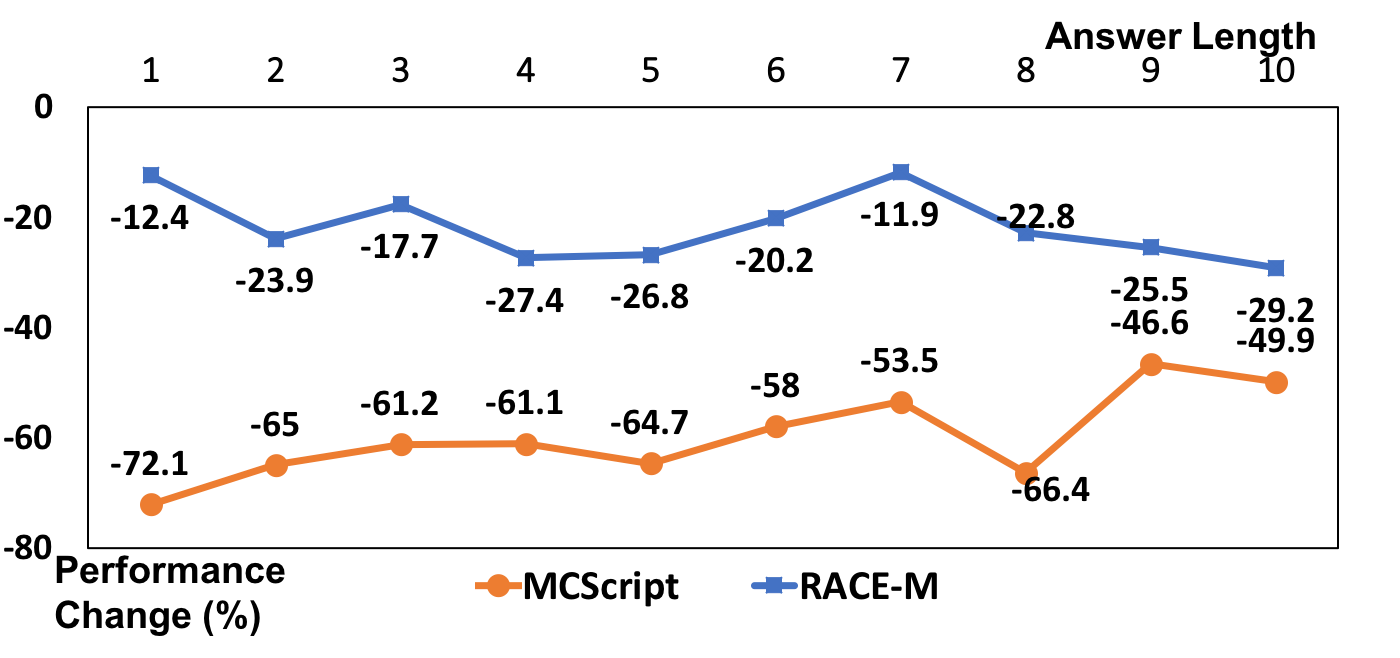}
    \caption{Plot of percentage performance change against answer length under AddSent2Opt-Shuffle Attack. Correlation Coefficient: MCScript: 0.7693, RACE-M: -0.4048.}
    \label{drop-anslen}
\end{figure}

\paragraph{Whether shorter answers are easier to attack}
We plot the percentage performance changes against the correct answer length, as shown in Figure~\ref{drop-anslen}. We do not observe strong correlation between the performance of our attaching method and the answer length.

\begin{figure}[h]
    \centering
    \includegraphics[width=7.3cm]{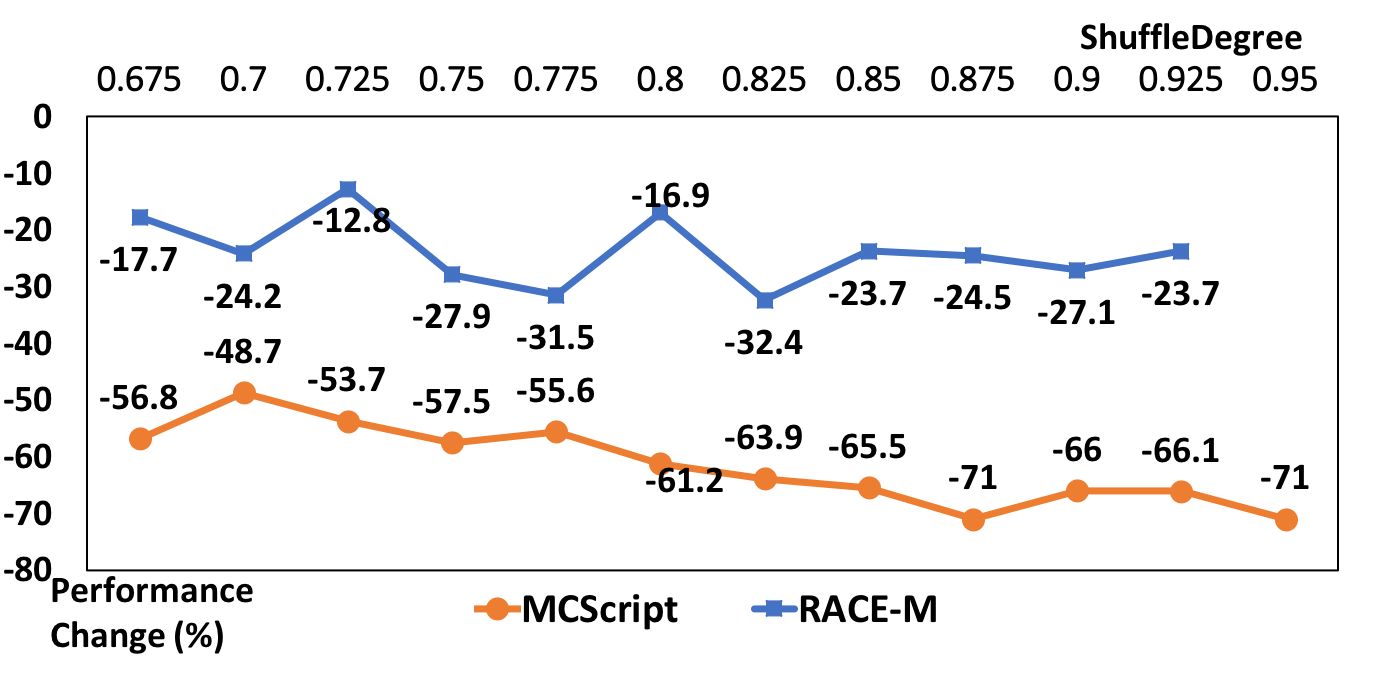}
    \caption{Plot of percentage performance change against ShuffleDegree under AddSent2Opt-Shuffle Attack. Correlation Coefficient: MCScript: -0.8965, RACE-M: -0.3442.}
    \label{random_plot_addsent2opt}
\end{figure}

\paragraph{Effect of random shuffling} 
For our attacking method ``AddSent2Opt-Shuffle", we randomly shuffle the sequence as our attacking options.
We will analyse how the degree of shuffling affects the performance. 
We plot the accuracy of BERT within different \textit{ShuffleDegree} ranges, which reflects the difference between shuffled sequence and the original sequence,  in Figure~\ref{random_plot_addsent2opt}.  Note that the \textit{ShuffleDegree} in the plot represents the largest \textit{ShuffleDegree} of all the shuffled options of each question. 
We can observe a weak tendency that with higher \textit{ShuffleDegree} the performance of BERT drops more.

\begin{figure}[h]

      \small
      \framebox{\parbox{7.5cm}{
      \textbf{Source: MCScript2.0}
      
      \textbf{Passage:} 
     I saw a sign for large pizza for 5.99 at domino pizza.
     I wanted to invite some people to my house. I made a list of toppings that all like. \textbf{\textit{(Randomly Selected Sentence:)} Then, I found a number to call.} I told him carry out. Then he asked me about the type of crust and I told him hand toast.  I ordered all large pizzas. 2 of them just plain cheese, 2 of them with beef and chicken, 2 of them with jalapeno and green pepper. 

      \textbf{Question:} 
    When did they bring total \$38.25?
        
      \textbf{Original Options:} 
      
      
    A \textbf{[Answer]}. when pizza delivered
    
    B. at the counter
      
      \textbf{New Distractor B after AddSent2Opt-Shuffle:}
    
    the . , Then I a number counter at found to call 
      
    \textbf{Probability Change:}
    
\begin{tabular}{lcc }
\toprule
Probability & Answer & Distractor
 \\
Empty Distractor & 0.893 & 0.107 \\
+ the  & 0.880 & 0.120 \\
+ . & 0.822 &  0.178 \\
+ , & 0.808 & 0.192 \\
+ Then  & 0.782 & 0.218 \\
+ I & 0.798 &  0.202 \\
+ a & 0.794 &  0.206 \\
+ number &  0.458 & 0.542 \\
+ counter & 0.583 &  0.417 \\
+ at &  0.496 & 0.504 \\
+ found & 0.265 & 0.735 \\
+ to & 0.216 &  0.784 \\
+ call  & 0.099 & 0.901 \\
 
 \bottomrule
\end{tabular}

      
      
      
      
      
      
      
      
      
      
      
      
      
      
      
      
      
      
      
      
}}
      \caption{Case Study of AddSent2Opt-Shuffle, we initialize the distractor as a empty sentence paddings and add the token from the AddSent2Opt-Shuffle distractor one by one to analyse the predicted probability change. The correct answer remains unchanged in this process. 
      }
      \label{case_study}
   \end{figure}

\paragraph{Case Study}

In Figure \ref{case_study}, we show a case study of how the predicted probabilities of the correct answer and distractor options change when each word of the distracting sequence is added to the distractor option. We can observe that: 1) Keyword matching plays an important role, for example, adding the word ``number'' to the distratcor option that matches the number ``38.25'' in the question significantly decreases the probability of the correct answer. 2) Keywords from the original distractor can also help the model identify the correct answer. For example, adding the word ``counter'' from the original distractor increases the probability for the correct answer. 3) Adding more matched words to the distratcor options can decrease the probability for the correct answer. 4) Stopwords can also influence the prediction.

\section{Conclusion}
In this work, we explore what BERT learns from MCRC datasets through un-readable data attack and un-answerable data training.
We proposed 3 different methods to attack BERT, and find that when un-readable distracting information is added to either the passage or the original distractors, BERT is highly likely to be fooled.
In this way, we show that BERT relies heavily on the keywords to solve multi-choice reading comprehension tasks.
We also use randomly shuffled input and partial input to train BERT, and observe that BERT could still learn surprisingly well how to answer the questions correctly.
This shows that BERT does not need the original correct syntactic and semantic information from the datasets to solve the task. In particular, the high performance on partial training shows that BERT can exploit the dataset artifacts and statistical cues to perform well instead of learning natural language understading and reasoning.
To make the model better understand natural language, both the datasets and the model need to be further improved.

\bibliography{acl2020}

\begin{thebibliography}{29}
\expandafter\ifx\csname natexlab\endcsname\relax\def\natexlab#1{#1}\fi

\bibitem[{Alzantot et~al.(2018)Alzantot, Sharma, Elgohary, Ho, Srivastava, and
  Chang}]{GenAdv}
Moustafa Alzantot, Yash Sharma, Ahmed Elgohary, Bo-Jhang Ho, Mani~B.
  Srivastava, and Kai-Wei Chang. 2018.
\newblock Generating natural language adversarial examples.
\newblock In \emph{EMNLP}.

\bibitem[{Clark et~al.(2019)Clark, Khandelwal, Levy, and Manning}]{BERTattn}
Kevin Clark, Urvashi Khandelwal, Omer Levy, and Christopher~D. Manning. 2019.
\newblock What does bert look at? an analysis of bert's attention.
\newblock \emph{ArXiv}, abs/1906.04341.

\bibitem[{Devlin et~al.(2019)Devlin, Chang, Lee, and Toutanova}]{BERT}
Jacob Devlin, Ming-Wei Chang, Kenton Lee, and Kristina Toutanova. 2019.
\newblock Bert: Pre-training of deep bidirectional transformers for language
  understanding.
\newblock In \emph{NAACL}.

\bibitem[{Ebrahimi et~al.(2018)Ebrahimi, Rao, Lowd, and Dou}]{hotflip}
Javid Ebrahimi, Anyi Rao, Daniel Lowd, and Dejing Dou. 2018.
\newblock Hotflip: White-box adversarial examples for text classification.
\newblock In \emph{ACL}.

\bibitem[{Goldberg(2019)}]{BERTsyntax}
Yoav Goldberg. 2019.
\newblock Assessing bert's syntactic abilities.
\newblock \emph{ArXiv}, abs/1901.05287.

\bibitem[{Gururangan et~al.(2018)Gururangan, Swayamdipta, Levy, Schwartz,
  Bowman, and Smith}]{artifacts}
Suchin Gururangan, Swabha Swayamdipta, Omer Levy, Roy Schwartz, Samuel~R.
  Bowman, and Noah~A. Smith. 2018.
\newblock Annotation artifacts in natural language inference data.
\newblock In \emph{NAACL-HLT}.

\bibitem[{Iyyer et~al.(2018)Iyyer, Wieting, Gimpel, and Zettlemoyer}]{SCPN}
Mohit Iyyer, John Wieting, Kevin Gimpel, and Luke~S. Zettlemoyer. 2018.
\newblock Adversarial example generation with syntactically controlled
  paraphrase networks.
\newblock In \emph{NAACL-HLT}.

\bibitem[{Jia and Liang(2017)}]{AdvSquad}
Robin Jia and Percy~S. Liang. 2017.
\newblock Adversarial examples for evaluating reading comprehension systems.
\newblock In \emph{EMNLP}.

\bibitem[{Kaushik and Lipton(2018)}]{reading}
Divyansh Kaushik and Zachary~Chase Lipton. 2018.
\newblock How much reading does reading comprehension require? a critical
  investigation of popular benchmarks.
\newblock In \emph{EMNLP}.

\bibitem[{Lai et~al.(2017)Lai, Xie, Liu, Yang, and Hovy}]{RACE}
Guokun Lai, Qizhe Xie, Hanxiao Liu, Yiming Yang, and Eduard~H. Hovy. 2017.
\newblock Race: Large-scale reading comprehension dataset from examinations.
\newblock In \emph{EMNLP}.

\bibitem[{Lan et~al.(2019)Lan, Chen, Goodman, Gimpel, Sharma, and
  Soricut}]{ALBERT}
Zhenzhong Lan, Mingda Chen, Sebastian Goodman, Kevin Gimpel, Piyush Sharma, and
  Radu Soricut. 2019.
\newblock Albert: A lite bert for self-supervised learning of language
  representations.
\newblock \emph{ArXiv}, abs/1909.11942.

\bibitem[{Liu et~al.(2019{\natexlab{a}})Liu, Gardner, Belinkov, Peters, and
  Smith}]{transferability}
Nelson~F. Liu, Matt Gardner, Yonatan Belinkov, Matthew~E. Peters, and Noah~A.
  Smith. 2019{\natexlab{a}}.
\newblock Linguistic knowledge and transferability of contextual
  representations.
\newblock In \emph{NAACL-HLT}.

\bibitem[{Liu et~al.(2019{\natexlab{b}})Liu, Ott, Goyal, Du, Joshi, Chen, Levy,
  Lewis, Zettlemoyer, and Stoyanov}]{RoBERTa}
Yinhan Liu, Myle Ott, Naman Goyal, Jingfei Du, Mandar Joshi, Danqi Chen, Omer
  Levy, Mike Lewis, Luke~S. Zettlemoyer, and Veselin Stoyanov.
  2019{\natexlab{b}}.
\newblock Roberta: A robustly optimized bert pretraining approach.
\newblock \emph{ArXiv}, abs/1907.11692.

\bibitem[{Niven and Kao(2019)}]{statsCue}
Timothy Niven and Hung-Yu Kao. 2019.
\newblock Probing neural network comprehension of natural language arguments.
\newblock In \emph{ACL}.

\bibitem[{Ostermann et~al.(2018)Ostermann, Modi, Roth, Thater, and
  Pinkal}]{MCScript}
Simon Ostermann, Ashutosh Modi, Michael~A. Roth, Stefan Thater, and Manfred
  Pinkal. 2018.
\newblock Mcscript: A novel dataset for assessing machine comprehension using
  script knowledge.
\newblock \emph{CoRR}, abs/1803.05223.

\bibitem[{Ostermann et~al.(2019)Ostermann, Roth, and Pinkal}]{mcscript2}
Simon Ostermann, Michael Roth, and Manfred Pinkal. 2019.
\newblock Mcscript2.0: A machine comprehension corpus focused on script events
  and participants.
\newblock In \emph{*SEM}.

\bibitem[{Peters et~al.(2018)Peters, Neumann, Iyyer, Gardner, Clark, Lee, and
  Zettlemoyer}]{ELMO}
Matthew~E. Peters, Mark Neumann, Mohit Iyyer, Matt Gardner, Christopher Clark,
  Kenton Lee, and Luke~S. Zettlemoyer. 2018.
\newblock Deep contextualized word representations.
\newblock In \emph{NAACL}.

\bibitem[{Peters et~al.(2019)Peters, Ruder, and Smith}]{tuneOrNot}
Matthew~E. Peters, Sebastian Ruder, and Noah~A. Smith. 2019.
\newblock To tune or not to tune? adapting pretrained representations to
  diverse tasks.
\newblock In \emph{RepL4NLP@ACL}.

\bibitem[{Poliak et~al.(2018)Poliak, Naradowsky, Haldar, Rudinger, and
  Durme}]{hypoBasline}
Adam Poliak, Jason Naradowsky, Aparajita Haldar, Rachel Rudinger, and
  Benjamin~Van Durme. 2018.
\newblock Hypothesis only baselines in natural language inference.
\newblock In \emph{*SEM@NAACL-HLT}.

\bibitem[{Radford et~al.(2018)Radford, Narasimhan, Salimans, and
  Sutskever}]{GPT}
Alec Radford, Karthik Narasimhan, Tim Salimans, and Ilya Sutskever. 2018.
\newblock Improving language understanding by generative pre-training.

\bibitem[{Ribeiro et~al.(2018)Ribeiro, Singh, and Guestrin}]{SEA}
Marco~Tulio Ribeiro, Sameer Singh, and Carlos Guestrin. 2018.
\newblock Semantically equivalent adversarial rules for debugging nlp models.
\newblock In \emph{ACL}.

\bibitem[{Richardson et~al.(2013)Richardson, Burges, and Renshaw}]{MCTest}
Matthew Richardson, Christopher J.~C. Burges, and Erin Renshaw. 2013.
\newblock Mctest: A challenge dataset for the open-domain machine comprehension
  of text.
\newblock In \emph{EMNLP}.

\bibitem[{Sankar et~al.(2019)Sankar, Subramanian, Pal, Chandar, and
  Bengio}]{shuffleDialog}
Chinnadhurai Sankar, Sandeep Subramanian, Christopher~Joseph Pal, Sarath
  Chandar, and Yoshua Bengio. 2019.
\newblock Do neural dialog systems use the conversation history effectively? an
  empirical study.
\newblock In \emph{ACL}.

\bibitem[{Sun et~al.(2019)Sun, Yu, Chen, Yu, Choi, and Cardie}]{DREAM}
Kai Sun, Dian Yu, Jianshu Chen, Dong Yu, Yejin Choi, and Claire Cardie. 2019.
\newblock Dream: A challenge data set and models for dialogue-based reading
  comprehension.
\newblock \emph{Transactions of the Association for Computational Linguistics},
  7:217--231.

\bibitem[{Tenney et~al.(2019)Tenney, Das, and Pavlick}]{BERTpipeline}
Ian Tenney, Dipanjan Das, and Ellie Pavlick. 2019.
\newblock Bert rediscovers the classical nlp pipeline.
\newblock In \emph{ACL}.

\bibitem[{Vaswani et~al.(2017)Vaswani, Shazeer, Parmar, Uszkoreit, Jones,
  Gomez, Kaiser, and Polosukhin}]{transformer}
Ashish Vaswani, Noam Shazeer, Niki Parmar, Jakob Uszkoreit, Llion Jones,
  Aidan~N. Gomez, Lukasz Kaiser, and Illia Polosukhin. 2017.
\newblock Attention is all you need.
\newblock In \emph{NIPS}.

\bibitem[{Wallace et~al.(2019)Wallace, Feng, Kandpal, Gardner, and
  Singh}]{universalTrigger}
Eric Wallace, Shi Feng, Nikhil Kandpal, Matt Gardner, and Sameer Singh. 2019.
\newblock Universal adversarial triggers for attacking and analyzing nlp.
\newblock In \emph{EMNLP}.

\bibitem[{Warstadt et~al.(2019)Warstadt, Cao, Grosu, Peng, Blix, Nie, Alsop,
  Bordia, Liu, Parrish, Wang, Phang, Mohananey, Htut, Jeretic, and
  Bowman}]{BERT-NPI}
Alex Warstadt, Yu~Cao, Ioana Grosu, Wei Peng, Hagen Blix, Yining Nie, Anna
  Alsop, Shikha Bordia, Haokun Liu, Alicia Parrish, Sheng-Fu Wang, Jason Phang,
  Anhad Mohananey, Phu~Mon Htut, Paloma Jeretic, and Samuel~R. Bowman. 2019.
\newblock Investigating bert's knowledge of language: Five analysis methods
  with npis.
\newblock In \emph{EMNLP}.

\bibitem[{Yang et~al.(2019)Yang, Dai, Yang, Carbonell, Salakhutdinov, and
  Le}]{xlnet}
Zhilin Yang, Zihang Dai, Yiming Yang, Jaime~G. Carbonell, Ruslan Salakhutdinov,
  and Quoc~V. Le. 2019.
\newblock Xlnet: Generalized autoregressive pretraining for language
  understanding.
\newblock \emph{ArXiv}, abs/1906.08237.

\end{thebibliography}
\bibliographystyle{acl_natbib}




\end{document}